\documentclass[letterpaper, 10 pt, conference]{ieeeconf}  

\IEEEoverridecommandlockouts                              

\usepackage{multirow}

\title{\LARGE \bf
The TUM VI Benchmark for Evaluating Visual-Inertial Odometry
}

\usepackage{tikz}
\usetikzlibrary{calc}
\usepackage{amsmath}
\usepackage{amssymb}
\usepackage{siunitx}
\usepackage{makecell}
\usepackage{hyperref}
\usepackage{cleveref}
\usepackage{pgfplots}
\usepgfplotslibrary{units}
\usepackage{bm}

\usepackage{balance}

\usepackage{booktabs,color}
\usepackage{tabularx}

\usepackage{dirtree}

\usepackage{array}
\usepackage{ragged2e}
\newcolumntype{P}[1]{>{\RaggedRight\hspace{0pt}}p{#1}}

\DeclareMathOperator{\trans}{trans}
\newcommand{\SE}{\mathrm{SE}}
\renewcommand{\vec}[1]{\mathbf{#1}}

\author{David Schubert$^*$, Thore Goll$^*$, Nikolaus Demmel$^*$, Vladyslav Usenko$^*$, J\"org St\"uckler and Daniel Cremers
\thanks{$^{*}$ These authors contributed equally. The authors are with Technical University of Munich, 85748 Garching bei M\"unchen, Germany
{\tt\small \{schubdav, gollt, demmeln, usenko, stueckle, cremers\}@in.tum.de}}%
}

\begin{document}

\newcommand{\point}[1]{\mathbf{#1}}

\maketitle
\thispagestyle{empty}
\pagestyle{empty}

\begin{abstract}
Visual odometry and SLAM methods have a large variety of applications in domains such as augmented reality or robotics. Complementing vision sensors with inertial measurements tremendously improves tracking accuracy and robustness, and thus has spawned large interest in the development of visual-inertial (VI) odometry approaches. 
In this paper, we propose the TUM VI benchmark, a novel dataset with a diverse set of sequences in different scenes for evaluating VI odometry. It provides camera images with 1024x1024 resolution at 20\,Hz, high dynamic range and photometric calibration. An IMU measures accelerations and angular velocities on 3~axes at 200\,Hz, while the cameras and IMU sensors are time-synchronized in hardware.
For trajectory evaluation, we also provide accurate pose ground truth from a motion capture system at high frequency (120\,Hz) at the start and end of the sequences which we accurately aligned with the camera and IMU measurements.
The full dataset with raw and calibrated data is publicly available. We also evaluate state-of-the-art VI odometry approaches on our dataset.
\end{abstract}

\section{INTRODUCTION}

Visual odometry and SLAM is a very active field of research with an abundance of applications in fields such as augmented reality or robotics. Variants include monocular (\cite{jin2000_rtsfm,engel2018_dso}), stereo (\cite{wang2017_stereodso,olson_stereovo}) and visual-inertial (\cite{bloesch2017_rovio,leutenegger15,usenko16icra}) methods. Compared to one camera, adding a second one in a stereo setup provides better robustness and scale-observability. Adding an inertial measurement unit (IMU) helps dealing with untextured environments and rapid motions and makes roll and pitch directly observable. On the other hand, the camera complements the IMU with external referencing to the environment in 6 degrees of freedom.

To compare competing methods, it is necessary to have publicly available data with ground truth. Given the relevance of the topic of visual-inertial odometry, the availability of high-quality datasets is surprisingly small. Compared to single-camera, purely visual datasets, the challenge with a stereo visual-intertial dataset lies in the accurate synchronization of three sensors. A commonly used option for evaluating visual-inertial odometry is the EuRoC MAV dataset~\cite{burri2016_eurocmavdataset}, but its image resolution and bit depth is not quite state-of-the-art anymore, and the number and variability of scenes is very limited.

For direct methods, which do not align pixel coordinates but image intensities, the assumption that the same 3D point has the same intensity in two different images should be satisfied. It has been shown that providing a photometric calibration that allows to compensate for exposure times, camera response function and lense vignetting is beneficial in this case~\cite{engel2018_dso}, however it is not a common feature of existing datasets.

In this paper, we propose the {\bf TUM VI benchmark}, a novel dataset with a diverse set of sequences in different scenes, with 1024x1024 image resolution at \SI{20}{\hertz}, 16-bit color depth, known exposure times, linear response function and vignette calibration. An IMU provides 3-axis accelerometer and gyro measurements at \SI{200}{\hertz}, which we correct for axis scaling and misalignment, while the cameras and IMU sensors are time-synchronized in hardware.
We recorded accurate pose ground truth with a motion capture system at high frequency (\SI{120}{\hertz}) which is available at the start and end of the sequences.
For accurate alignment of sensor measurements with the ground truth, we calibrated time offsets and relative transforms.

We evaluate state-of-the-art visual-inertial algorithms on our dataset.
The full dataset with raw and calibrated data, together with preview videos, is available on:
\begin{center}
\color{purple}
\urlstyle{tt}
\textbf{\url{https://vision.in.tum.de/data/datasets/visual-inertial-dataset}}
\end{center}

\begin{figure}
    \centering
    \includegraphics[width=\linewidth]{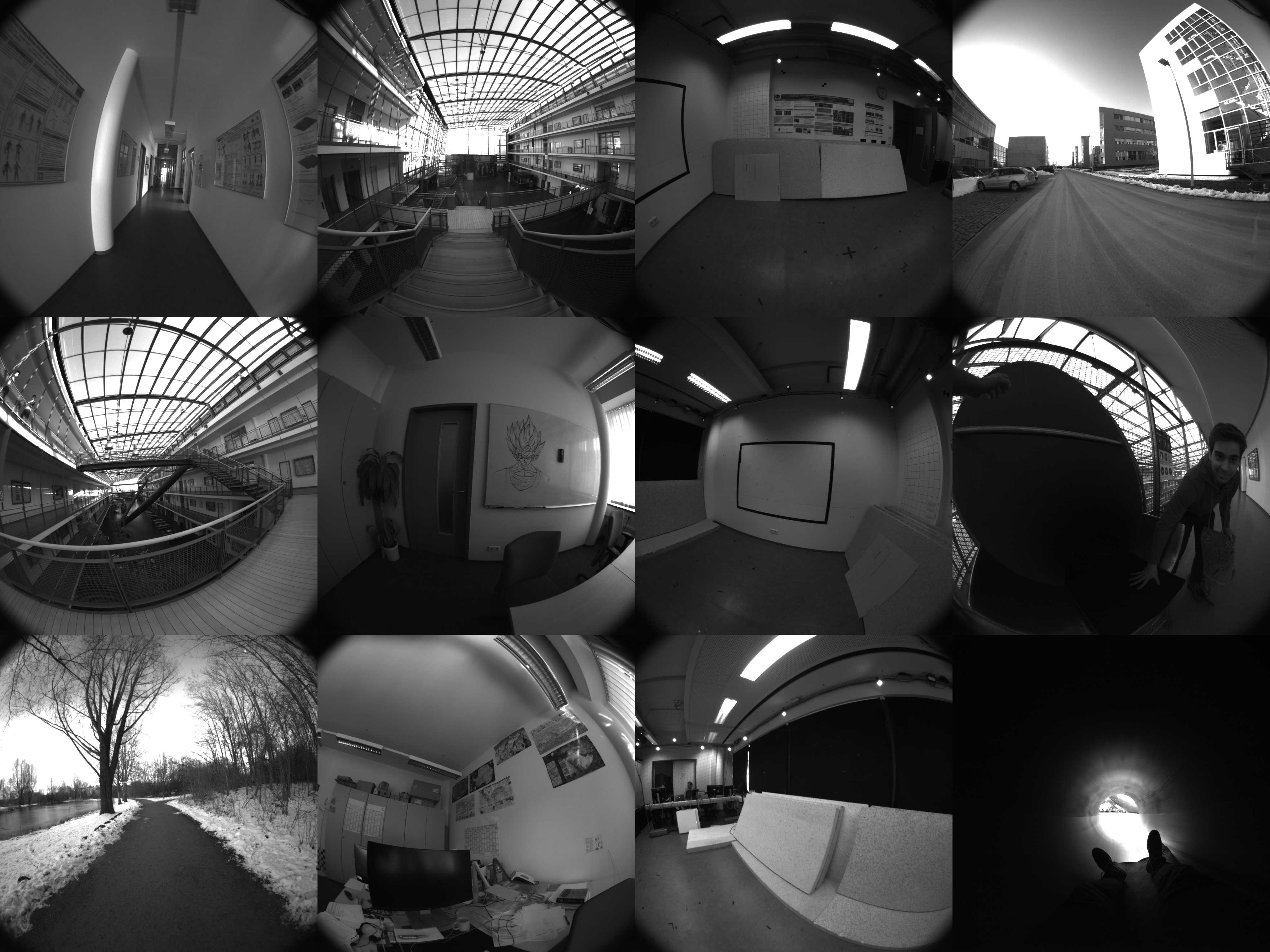}
    \caption{The \textbf{TUM VI benchmark} includes synchronized measurements from an IMU and a stereo camera in several challenging indoor and outdoor sequences. The cameras are equipped with large field-of-view lenses (195$^\circ$) and provide high dynamic range images (16 bit) at high resolution (1 MP) with linear response function. The figure shows example frames from the dataset.}
    \label{fig:thumbs}
\end{figure}

\section{RELATED WORK}

\begin{table*}
\caption{Comparison of datasets with vision and imu data.}
\label{tab:relateddatasets}
\centering
\bgroup
\def\arraystretch{1.4}
\begin{tabularx}{0.99\textwidth}{P{2cm}lp{1.2cm}p{1cm}P{2.2cm}P{2cm}P{0.8cm}P{2.7cm}P{1.5cm}}
\toprule
dataset & year  & environ. & carrier & cameras & IMUs & time sync & ground truth & stats/props \\
\midrule
Kitti Odometry~\cite{geiger2013_kitti} & 2013 & outdoors & car & 1~{\bf stereo} RGB 2x1392x512 @10Hz, 1~stereo gray 2x1392x512 @10Hz  & OXTS RT 3003 3-axis acc/gyro @10Hz & sw & OXTS RT 3003 pose @10Hz, acc. $<$10cm & 22~seqs, 39.2\,km\\

Malaga Urban~\cite{blanco2014_malagadataset} & 2014 & outdoors & car & 1~{\bf stereo} RGB 2x1024x768 @20Hz & 3-axis acc/gyro @100Hz & sw & GPS pos @1Hz, low acc & 15~subseqs, 36.8\,km \\

UMich NCLT~\cite{carkevaris2015_umichnclvdataset} & 2015 & \bf in-/outdoors & Segway & 6~RGB (omni) 1600x1200 @5Hz & 3-axis acc/gyro @100Hz & sw & fused GPS/IMU/laser pose @150Hz, acc$\approx$10cm  & 27~seqs, 147.3\,km \\

EuRoC MAV~\cite{burri2016_eurocmavdataset} & 2016 & indoors & MAV & 1~{\bf stereo} gray 2x752x480 @20Hz & ADIS16488 3-axis acc/gyro @200Hz & \bf hw & laser tracker pos @20Hz, {\bf motion capture pose @100Hz}, acc$\approx$1mm & 11~seqs, 0.9\,km\\ 


PennCOSYVIO \cite{pfrommer2017_penncosyvio} & 2017 & \bf in-/outdoors & handheld & 4~RGB 1920x1080 @30Hz (rolling shutter), 1~{\bf stereo} gray 2x752x480 @20Hz, 1~fisheye gray 640x480 @30Hz & ADIS16488 3-axis acc/gyro @200Hz, Tango 3-axis acc @128Hz / 3-axis gyro @100Hz & {\bf hw} (stereo gray/ ADIS), sw & fiducial markers pose@30Hz, acc$\approx$15cm & 4~seqs, 0.6\,km \\

Zurich Urban MAV~\cite{majdik2017_urbanmavdataset} & 2017 & outdoors & MAV & 1~RGB 1920x1080 @30Hz (rolling shutter) & 3-axis acc/gyro @10Hz & sw & Pix4D visual pose, acc unknown & 1~seq, 2\,km\\

\bf Ours (TUM VI) & 2018 & \bf in-/outdoors & handheld & 1~{\bf stereo} gray 2x1024x1024 @20Hz & BMI160 3-axis acc/gyro @200Hz & \bf hw & {\bf partial motion capture pose @120Hz}, marker pos acc$\approx$1mm (static case) & 28~seqs, 20\,km, {\bf photometric calibration} \\
\bottomrule
\end{tabularx}
\egroup
\end{table*}

Datasets have in the past greatly fostered the research of visual odometry and SLAM algorithms.
In \cref{tab:relateddatasets} we give an overview over the most relevant datasets that include vision and IMU data.

{\bf Visual odometry and SLAM datasets:} The TUM \mbox{RGB-D} dataset~\cite{sturm2012benchmark} is focused on the evaluation of RGB-D odometry and SLAM algorithms and has been extensively used by the research community.
It provides 47 RGB-D sequences with ground-truth pose trajectories recorded with a motion capture system.
It also comes with evaluation tools for measuring drift and SLAM trajectory alignment. 
For evaluating monocular odometry, recently the TUM MonoVO dataset~\cite{engel2016photometrically} has been proposed.
The dataset contains 50 sequences in indoor and outdoor environments and has been photometrically calibrated for exposure times, lens vignetting and camera response function.
Drift can be assessed by comparing the start and end position of the trajectory which coincide for the recordings.
We also provide photometric calibration for our dataset, but additionally recorded motion capture ground truth in parts of the trajectories for better pose accuracy assessment. 
Furthermore, the above datasets do not include time-synchronized IMU measurements with the camera images like our benchmark.

For research on autonomous driving, visual odometry and SLAM datasets have been proposed such as Kitti~\cite{geiger2013_kitti}, Malaga Urban dataset~\cite{blanco2014_malagadataset}, or the Robot Oxford car dataset~\cite{maddern2017_oxfordrobotcardataset}. 
The Kitti and Malaga Urban datasets also include low-frequency IMU information which is, however, not time-synchronized with the camera images.
While Kitti provides a GPS/INS-based ground truth with accuracy below 10\,cm, the Malaga Urban dataset only includes a coarse position for reference from a low-cost GPS sensor.
Our dataset contains \SI{20}{\hertz} camera images and hardware time-synchronized 3-axis accelerometer and gyro measurements at \SI{200}{\hertz}.
Ground-truth poses are recorded at \SI{120}{\hertz} and are accurately time-aligned with the sensor measurements as well.

{\bf Visual-inertial odometry and SLAM datasets:} Similar to our benchmark, some recent datasets also provide time-synchronized IMU measurements with visual data and have been designed for the evaluation of visual-inertial (VI) odometry and SLAM approaches.
The EuRoC MAV dataset~\cite{burri2016_eurocmavdataset} includes 11 indoor sequences recorded with a Skybotix stereo VI sensor from a MAV.
Accurate ground truth (approx. 1mm) is recorded using a laser tracker or a motion capture system.
Compared to our benchmark, the sequences in EuRoC MAV are shorter and have less variety as they only contain recordings in one machine hall and one lab room.
Furthermore, EuRoC MAV does not include a photometric calibration which is important to benchmark direct methods.
Further datasets for visual-inertial SLAM are the PennCOSYVIO dataset~\cite{pfrommer2017_penncosyvio} and the Zurich Urban MAV dataset~\cite{majdik2017_urbanmavdataset}.
However, they do not contain photometric calibration and as accurate ground truth or time-synchronization of IMU and camera images like our benchmark (cf.~\cref{tab:relateddatasets}).

\section{SENSOR SETUP}

Our sensor setup consists of two monochrome cameras in a stereo setup and an IMU, see \cref{fig:setup}. The left figure shows a schematic view of all involved coordinate systems. We use the convention that a pose $\mathbf{T}_{BA}\in\SE(3)$ transforms point coordinates $\point{p}_{A}\in\mathbb{R}^3$ in system $A$ to coordinates in $B$ through $\point{p}_{B}=\mathbf{T}_{BA}\point{p}_{A}$.
For the coordinate systems, we use the following abbreviations,
\begin{description}
    \item[I] IMU
    \item[$\text{C}_0$] camera 0
    \item[$\text{C}_1$] camera 1
    \item[M] IR-reflective markers
    \item[G] grid of AprilTags
    \item[W] world frame (reference frame of MoCap system)
\end{description}

The IMU is rigidly connected to the two cameras and several IR-reflective markers which allow for pose tracking of the sensor setup by the motion capture (MoCap) system. For calibrating the camera intrinsics and the extrinsics of the sensor setup, we use a grid of AprilTags \cite{olson2010apriltag} which has a fixed pose in the MoCap reference (world) system. In the following, we briefly describe the hardware components. An overview is also given in \cref{tab:sensors}.

\begin{figure*}
    \centering
    \input{setup_schematic.tikz}\\
    \caption{Sensor setup. Left: Schematic view of the different coordinate systems. The rounded rectangle contains all components which are rigidly connected with the IMU coordinate system. A dotted line indicates a temporally changing relative pose when moving the sensor. Right: Photo of the sensor setup. It contains two cameras in a stereo setup, a microcontroller board with integrated IMU, a luminance sensor between the cameras and IR reflective markers. }
    \label{fig:setup}
\end{figure*}

\begin{table}
\caption{Overview of sensors in our setup.}
\label{tab:sensors}
\centering
\begin{tabular}{llll}
\toprule
Sensor & Type & Rate & Characteristics \\
\midrule
Cameras & \makecell[lt]{2 $\times$ IDS uEye \\ UI-3241LE-M-GL} & \SI{20}{\hertz} & \makecell[lt]{global shutter \\ 1024x1024 \\ 16-bit gray} \\
IMU & Bosch BMI160 & \SI{200}{\hertz} & \makecell[lt]{3D accelerometer \\ 3D gyroscope \\ temperature} \\
MoCap & OptiTrack Flex13 & \SI{120}{\hertz} & \makecell[lt]{6D Pose \\ infrared cameras} \\
Light sensor & TAOS TSL2561 & \SI{200}{Hz} & scalar luminance\\
\bottomrule
\end{tabular}
\end{table}

\subsection{Camera}

We use two uEye UI-3241LE-M-GL cameras by IDS. Each has a global shutter CMOS sensor which delivers 1024x1024 monochrome images. The whole intensity range of the sensor can be represented using 16-bit images, so applying a non-linear response function (usually used to increase the precision at a certain intensity range) is not required. The cameras operate at \SI{20}{\hertz} and are triggered synchronously by a Genuino 101 microcontroller.

The cameras are equipped with Lensagon BF2M2020S23 lenses by Lensation. These fisheye lenses have a field of view of \SI{195}{\degree} (diagonal), though our cameras record a slightly reduced field of view in horizontal and vertical directions due to the sensor size.

\subsection{Light Sensor}

We design our sensor setup to ensure the same exposure time of corresponding images for the two cameras. This way, both camera images have the same brightness for corresponding image points (which otherwise needs to be calibrated or estimated with the visual odometry). Furthermore, this also ensures the same center of the exposure time (which is used as the image timestamp) for two corresponding images and allows us to record accurate per-frame exposure times. 

\begin{figure}
    
    \includegraphics[width=0.95\linewidth]{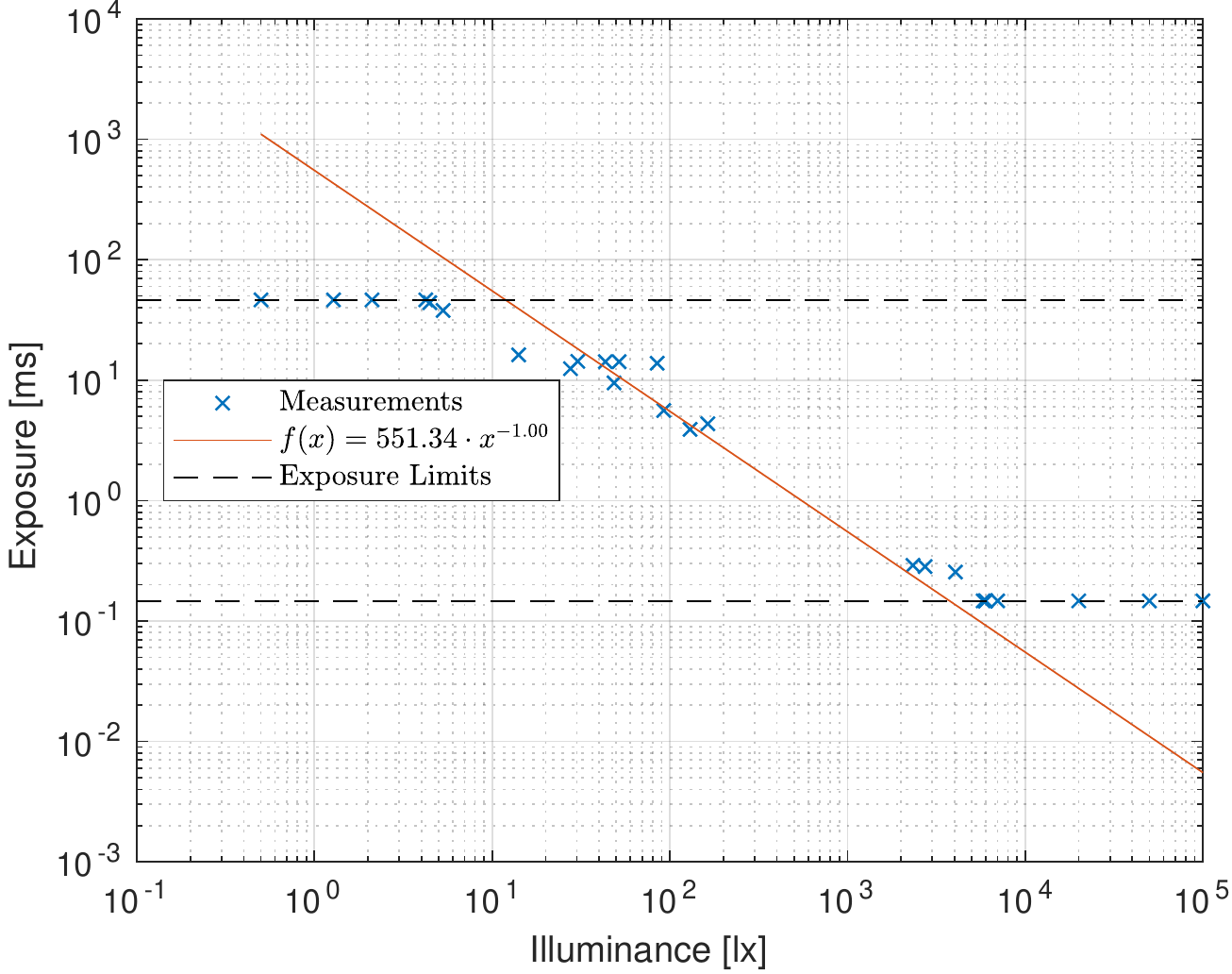}
    \caption{Relation of illuminance measurements by our light sensor and corresponding exposure time settings by the camera's auto exposure mode. The dashed lines show the minimum and maximum exposure times possible. The red line shows the least-squares fit (without saturated values) which we use for estimating the next exposure time.}
    
    \label{fig:luxsensor}
\end{figure}

We use a TSL2561 light sensor by TAOS to estimate the required exposure time. The sensor delivers an approximate measurement of the illuminance of the environment. The relation of these measurements and the exposure times which are selected by the camera's auto exposure is approximately inversely proportional, as can be seen in \cref{fig:luxsensor}. We find its parameters using a least-squares fit and use it to set the exposure times of both cameras based on the latest illuminance measurement. This assumes that the change in scene brightness between the light measurement and the start of the exposure is negligible. Note that it is not necessary to reproduce the cameras' auto exposure control exactly as long as too dark or too bright images can be avoided.
In most cases, the results of our exposure control approach are visually satisfying, but short video segments may be challenging.

\subsection{IMU}\label{ss:imu}

Our sensor setup includes a Bosch BMI160 IMU, which contains 16-bit 3-axis MEMS accelerometer and gyroscope. IMU temperature is recorded, facilitating temperature-dependent noise models. We set its output rate to \SI{200}{\hertz}. The IMU is integrated in the Genuino 101 microcontroller board which triggers the cameras and reads the IMU values. This way, the timestamps of cameras and IMU are well aligned. We estimate the remaining small constant time offset (owing to the readout delay of IMU measurements) during the camera-imu extrinsics calibration which yields a value of \SI{5.3}{ms} for our setup. We estimated this value once and corrected for it in both raw and calibrated datasets.

\subsection{Motion Capture System}

For recording accurate ground-truth poses at a high frame-rate of \SI{120}{\hertz}, we use an OptiTrack motion capture system. It consists of 16 infrared Flex13 cameras which track the IR-reflective markers on the sensor setup. The MoCap system only covers a single room, so we cannot record ground truth for parts of the longer trajectories outside the room. Instead, all sequences start and end in the MoCap room such that our sequences provide ground truth at the beginning and the end. 

\section{CALIBRATION}

We include two types of sensor data in our dataset: raw data and calibrated data. The raw data is measured directly by the sensors as described so far, but cannot be used without proper calibration. In the following, we describe which calibrations we apply to the raw data in order to make it usable.

\subsection{Camera Calibration}

Firstly, we calibrate the camera intrinsics and the extrinsics of the stereo setup.
We use one of the calib-cam sequences, where we took care to slowly move the cameras  in front of the calibration grid to keep motion blur as small as possible. 

\subsection{IMU and Hand-Eye Calibration}

We then calibrate the extrinsics between IMU and cameras as well as between IMU and MoCap frame.
Concurrently, we estimate the time-synchronization of IMU with MoCap measurements and IMU parameters such as axis alignment, scale differences and biases.

Specifically, we keep the camera intrinsics from the previous calibration fixed and optimize for
\begin{itemize}
    \item the relative pose between cameras and IMU,
    \item the time shift between MoCap and IMU time,
    \item the time shift between camera and IMU time,
    \item the relative pose between the cameras,
    \item the relative poses $\mathbf{T}_\text{MI}$ and $\mathbf{T}_\text{WG}$,
    \item coarse initial accelerometer and gyroscope biases $\mathbf{b}_\text{a}$ and $\mathbf{b}_\text{g}$,
    \item axis scaling and misalignment matrices as in \cite{rehder16} $\mathbf{M}_\text{a},\mathbf{M}_\text{g}\in\mathbb{R}^{3\times3}$.
\end{itemize}

The relative poses~$\mathbf{T}_\text{MI}$ and~$\mathbf{T}_\text{WG}$ are found through hand-eye calibration using a non-linear least squares fitting procedure. Using the relative poses, we convert raw MoCap poses $\mathbf{T}_\text{WM}$ to calibrated ground-truth poses $\mathbf{T}_\text{WI}$ for the IMU.

\begin{figure*}
\centering

\includegraphics[width=0.32\linewidth]{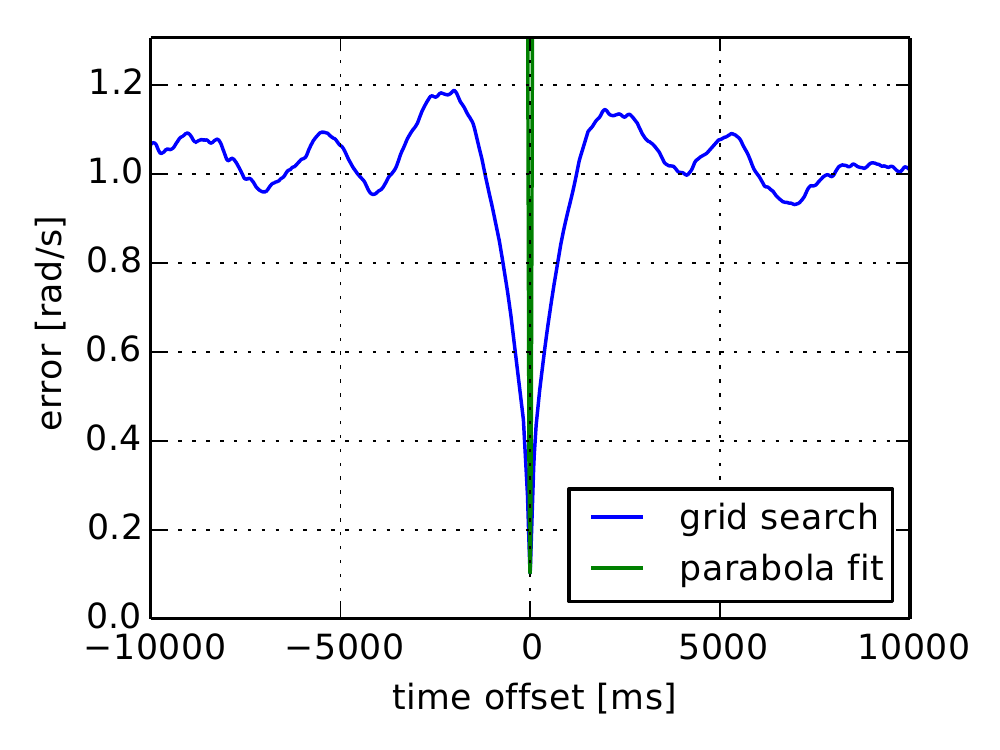}
\includegraphics[width=0.32\linewidth]{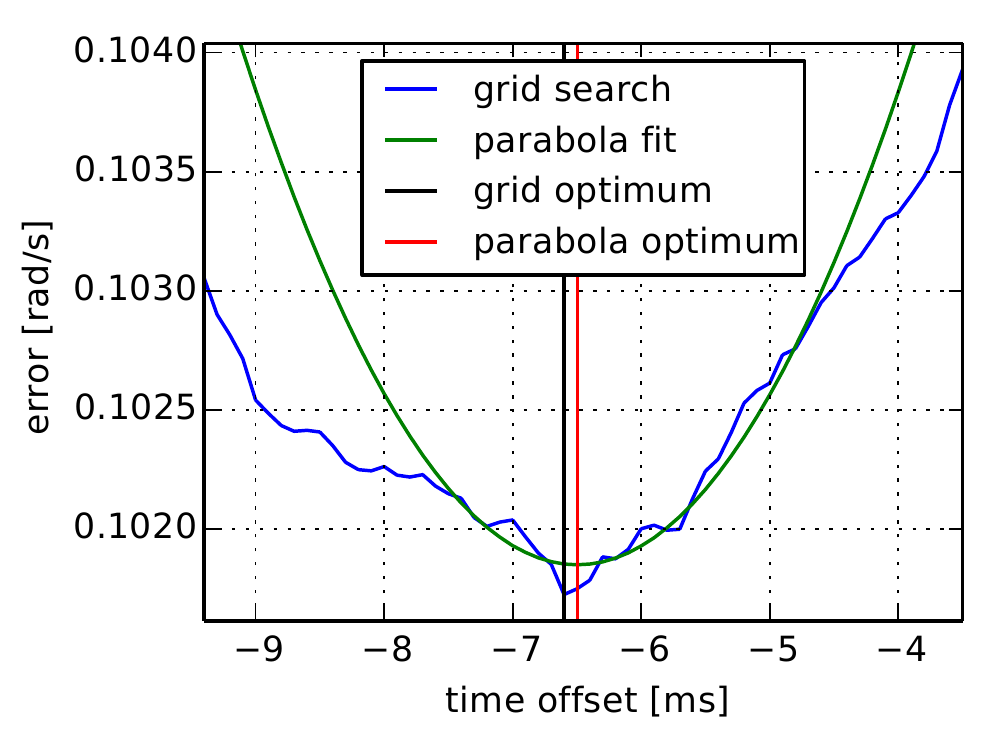}
\includegraphics[width=0.32\linewidth]{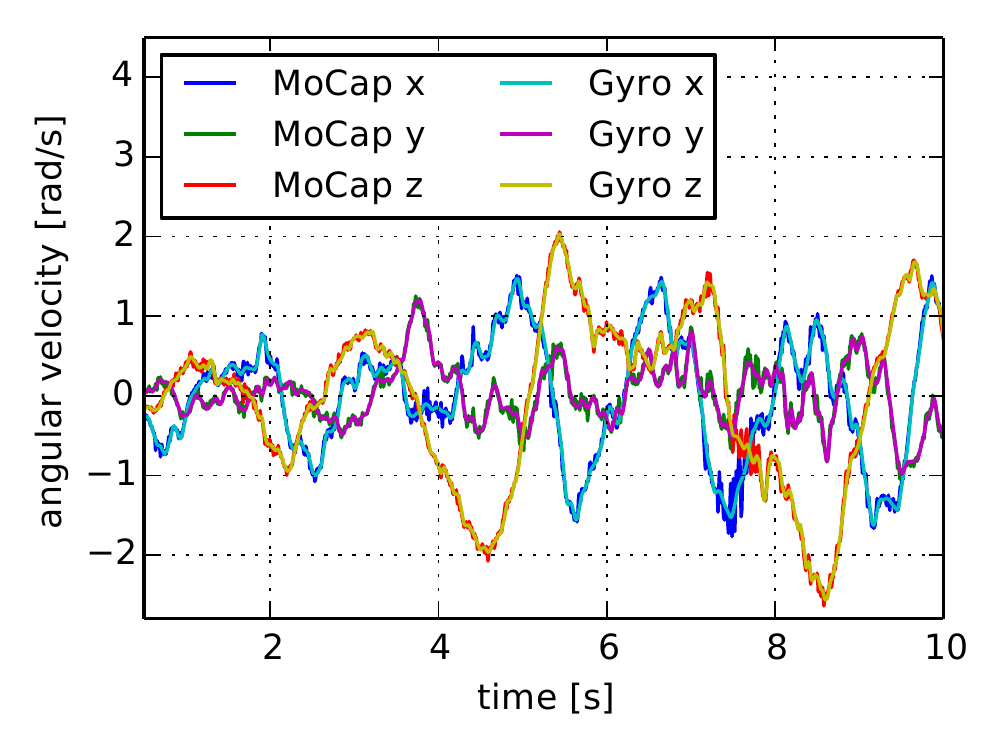}

\caption{Left and middle:
Time alignment is performed using grid search. After a coarse initialization it is followed by parabola fitting to find the sub-discretization minimum.
Right: Rotational velocities from gyroscope and MoCap after time alignment on the test sequence. MoCap angular velocities are computed using central differences on the orientation.}
\label{fig:timeshift}
\end{figure*}

Additionally, we compensate for the time shift between MoCap and IMU time in the calibrated data.
The time offset between MoCap and IMU has to be estimated for each sequence individually. To find the time offset, angular velocities are calculated from the MoCap poses and aligned with the gyroscope measurements. This is done --- after a coarse alignment based on measurement arrival time --- using a grid search with a stepsize of \SI{100}{\mu s}. Then a parabola is fitted around the minimum and the minimum of the parabola is the resulting time offset. The results of this procedure can be seen in \cref{fig:timeshift}. The ground-truth poses in the calibrated data are always given in IMU time.

We also compensate for axis/scale misalignment and initial biases of the raw accelerations~$\mathbf{a}_\text{raw}$ and angular velocities~$\boldsymbol{\omega}_\text{raw}$ using
\begin{align}
    \mathbf{a}_\text{calibrated} &= \mathbf{M}_\text{a} \cdot \mathbf{a}_\text{raw} - \mathbf{b}_\text{a}\,,\label{eq:aproc}\\
    \boldsymbol{\omega}_\text{calibrated} &= \mathbf{M}_\text{g} \cdot \boldsymbol{\omega}_\text{raw} - \mathbf{b}_\text{g}\,.\label{eq:wproc}
\end{align}
The matrices $\mathbf{M}_\text{a}, \mathbf{M}_\text{g}$ account for rotational misalignments of gyroscope and accelerometer, axes not being orthogonal or axes not having the same scale. For $\mathbf{M}_\text{g}$, all 9 entries are optimized, whereas $\mathbf{M}_\text{a}$ is chosen to be lower triangular with 6 parameters. The remaining three parameters (rotation) are redundant and have to be fixed in order to obtain a well-constrained system.

In principle, it is not necessary to deduct $\mathbf{b}_\text{a}$ and $\mathbf{b}_\text{g}$, as inertial state estimation algorithms usually estimate a time-varying bias. However, we found that in our hardware setup there is a large IMU bias that is coarsely reproducible between sensor restarts and therefore approximate precalibration is reasonable. Note that estimating the biases accurately from the sequences is still required for inertial state estimation.

For the calibration step, we use one of the calib-imu sequences which are recorded in front of the calibration grid with motions in all 6 degrees of freedom.

\subsection{IMU Noise Parameters}

\begin{figure*}
    \centering
    \includegraphics[width=\linewidth]{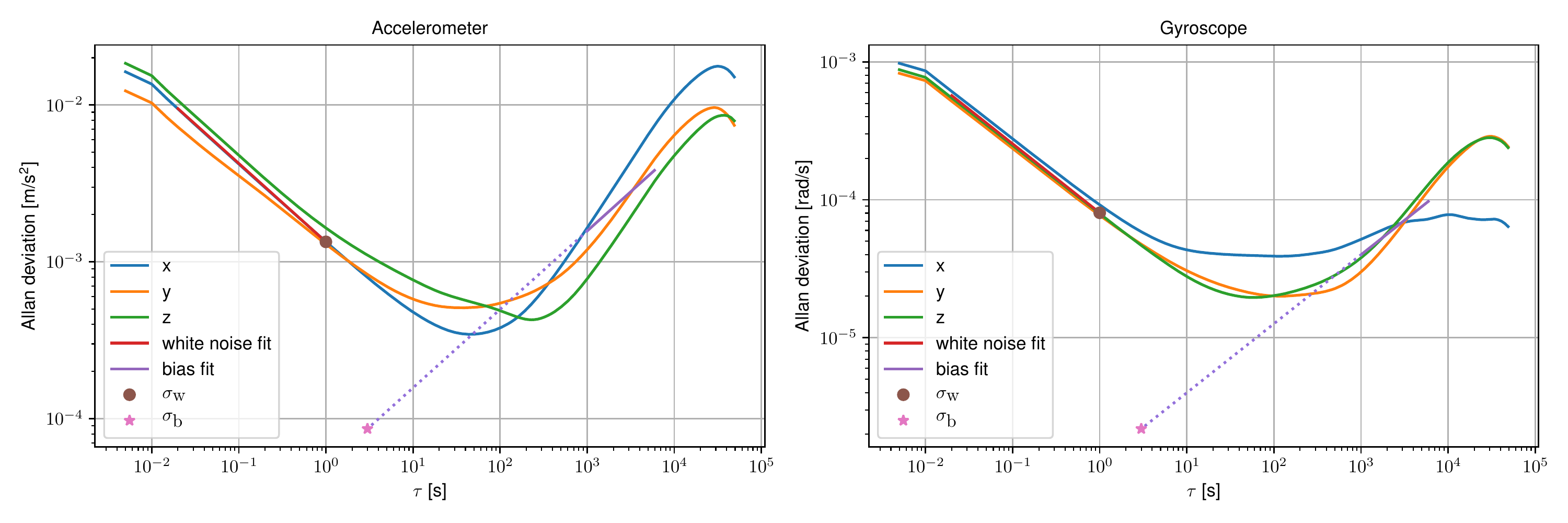}
    \caption{Allan deviation of both accelerometer (left) and gyroscope (right).
    For the fit with slope $-1/2$ we averaged over all three dimensions and took the range $0.02\leq\tau\leq 1$ into account. For the fit with slope $1/2$, the same averaging was done for the accelerometer, but for the gyroscope we only averaged the $y$-coordinate and the $z$-coordinate. The fit region is $\SI{1000}{}\leq\tau\leq\SI{6000}{}$. The assumed slope of 1/2 does not fit perfectly, which might be due to unmodeled effects such as temperature dependence.
    The numerical values of noise densities $\sigma_\text{w}$ can be found at an integration time of $\tau = 1s$ on the straight line with slope $-1/2$, while bias parameters $\sigma_\text{b}$ are identified as the value on the straight line with slope $1/2$ at an integration time of $\tau=3s$. This results in $\sigma_\text{w}=\SI{1.4e-3}{\metre\per\second\squared\per\sqrt{\hertz}}$,
    $\sigma_\text{b}=\SI{8.6e-05}{\metre\per\second\cubed\per\sqrt{\hertz}}$ for the accelerometer and
    $\sigma_\text{w}=\SI{8.0e-05}{\radian\per\second\per\sqrt{\hertz}}$,
    $\sigma_\text{b}=\SI{2.2e-06}{\radian\per\second\squared\per\sqrt{\hertz}}$ for the gyroscope.
    The white noise parameters are similar to typical values provided by the manufacturer,
    $\sigma_\text{w}=\SI{1.8e-3}{\metre\per\second\squared\per\sqrt{\hertz}}$ (accelerometer) and
    $\sigma_\text{w}=\SI{1.2e-4}{\radian\per\second\per\sqrt{\hertz}}$ (gyroscope).
}   
    \label{fig:allan}
\end{figure*}

For proper probabilistic modeling of IMU measurements in state estimation algorithms and accurate geometric calibration, the intrinsic noise parameters of the IMU are needed. We assume that our IMU measurements (accelerations or angular velocities) are perturbed by white noise with standard deviation $\sigma_\text{w}$ and a bias that is slowly changing according to a random walk, which is an integration of white noise with standard deviation $\sigma_\text{b}$. To estimate these quantities, we analyse their Allan deviation $\sigma_\text{Allan}(\tau)$ as a function of integration time $\tau$. For a resting IMU with only white noise present, the Allan deviation relates to the white noise standard deviation as

\begin{align}\label{eq:allan1}
    \sigma_\text{Allan}(\tau)=\frac{\sigma_\text{w}}{\sqrt{\tau}}\,,
\end{align}
so the numerical value of the parameter $\sigma_\text{w}$ can be found at $\tau=\SI{1}{s}$. If the measurement is only perturbed by the bias, the relation is
\begin{align}\label{eq:allan2}
    \sigma_\text{Allan}(\tau)=\sigma_\text{b}\sqrt{\frac{\tau}{3}}\,,
\end{align}
which means the parameter can be found at $\tau=\SI{3}{s}$. The relations between Allan deviation and integration time in Eqs.~\ref{eq:allan1} and \ref{eq:allan2} can be found in \cite{ieee1998imustandard} and we also provide a derivation in Appendix \ref{ap}, together with a definition and the estimation method for the Allan deviation.
White noise and bias dominate the Allan variance in different ranges of $\tau$. Thus, in the log-log plot of $\sigma_\text{Allan}(\tau)$ in \cref{fig:allan}, a straight line with slope $-\frac{1}{2}$ has been fitted to an appropriate range of the data to determine $\sigma_\text{w}$, and a straight line with slope $\frac{1}{2}$ has been fitted to another range to determine $\sigma_\text{b}$.

\subsection{Photometric Calibration}\label{sec:photocalib}

To enable good intensity matching for direct methods, we also provide vignette calibration. For this, we use the calibration code provided by the TUM MonoVO dataset\footnote{\url{https://github.com/tum-vision/mono_dataset_code}}~\cite{engel2016photometrically}. The image formation model is given by
\begin{align}
I(\vec{x}) = G\left(tV(\vec{x})B(\vec{x})\right)\,.
\end{align}
This means for an image point $\vec{x}$, light with intensity $B(\vec{x})$ is attenuated by a vignetting factor $V(\vec{x})\in[0,1]$, then is integrated during the exposure time $t$, and finally is converted by a response function $G$ into the irradiance value $I(\vec{x})$. In our case, we assume $G$ linear, so the model simplifies to $I(\vec{x})\propto tV(\vec{x})B(\vec{x})$. The given code requires images of a plane with a small calibration tag, taken from different viewpoints. It then alternatingly optimizes the texture of the wall (up to a constant factor) and a non-parametric vignette function. The result is a PNG image representing vignette values between 0 and 1 for each pixel.

\begin{table}
    \caption{RMSE RPE OF THE EVALUATED METHODS ON 1 SECOND SEGMENTS}
    \label{tab:rpe_results}
    \centering
    \begin{tabular}{lccc}
        \toprule
        Sequence & OKVIS & ROVIO & VINS \\
        \midrule
        room1 & \textbf{0.013m} / \textbf{0.43$^\circ$} & 0.029m / 0.53$^\circ$ & 0.015m / 0.44$^\circ$ \\
        room2 & \textbf{0.015m} / \textbf{0.62$^\circ$} & 0.030m / 0.67$^\circ$ & 0.017m / 0.63$^\circ$ \\
        room3 & \textbf{0.012m} / \textbf{0.63$^\circ$} & 0.027m / 0.66$^\circ$ & 0.023m / 0.63$^\circ$ \\
        room4 & \textbf{0.012m} / 0.57$^\circ$ & 0.022m / 0.61$^\circ$ & 0.015m / \textbf{0.41$^\circ$} \\
        room5 & \textbf{0.012m} / \textbf{0.47$^\circ$} & 0.031m / 0.60$^\circ$ & 0.026m / 0.47$^\circ$ \\
        room6 & \textbf{0.012m} / 0.49$^\circ$ & 0.019m / 0.50$^\circ$ & 0.014m / \textbf{0.44$^\circ$} \\
        \bottomrule
    \end{tabular}
\end{table}

\section{DATASET}

\subsection{Sequences}
Besides evaluation sequences, we also make our calibration data accessible such that users can perform their own calibration, even though we provide calibrated data and our calibration results. The sequences can be divided into the following categories.
\begin{itemize}
    \item \textbf{calib-cam:} for calibration of camera intrinsics and stereo extrinsics. A grid of AprilTags has been recorded at low frame rate with changing viewpoints and small camera motion.
    \item \textbf{calib-imu:} for cam-imu calibration to find the relative pose between cameras and IMU. Includes rapid motions in front of the April grid exciting all 6 degrees of freedom. A small exposure has been chosen to avoid motion blur.
    \item \textbf{calib-vignette:} for vignette calibration. Features motion in front of a white wall with a calibration tag in the middle.
    \item \textbf{imu-static:} only IMU data to estimate noise and random walk parameters (111 hours standing still).
    \item \textbf{room:} sequences completely inside the MoCap room such that the full trajectory is covered by the ground truth.
    \item \textbf{corridor:} sequences with camera motion along a corridor and to and from offices
    \item \textbf{magistrale:} sequences featuring a walk around the central hall in a university building
    \item \textbf{outdoors:} sequences of a larger walk outside on a university campus
    \item \textbf{slides:} sequences of a walk in the central hall of a university building including a small part sliding in a closed tube with no visual features.
\end{itemize}

\subsection{Format}

\subsubsection{ROS Bag Files}

For each sequence, we provide three different ROS bag files, one raw bag and two calibrated ones.
Raw bags contain the data as it has been recorded, i.e.\ before hand-eye, time shift or IMU calibration. They include the following topics.

\texttt{/cam0/image\_raw}
\par\texttt{/cam1/image\_raw}
\par\texttt{/imu0}
\par\texttt{/vrpn\_client/raw\_transform}

The first two contain the images of the cameras. Most fields in the messages are self-explanatory and follow standard conventions, but note that \texttt{frame\_id} provides the exposure time in nanoseconds. In the IMU topic, we do not give the orientation, but we use the second entry of \texttt{orientation\_covariance} to provide the temperature of the IMU in degree Celsius. The last topic contains the raw MoCap poses $\mathbf{T}_\text{WM}$. For each pose there is a timestamp in MoCap time, a translation vector and a rotation quaternion.

Calibrated bags contain the same topics as raw bags but with calibrated data. The differences are:
\begin{itemize}
\item MoCap poses have been aligned with the IMU frame (through hand-eye calibration, $\mathbf{T}_\text{WI}$),
\item outlier MoCap poses have been removed with a median filter on positions,
\item timestamps of the MoCap poses have been synchronized with the IMU time using the time shift calibration,
\item IMU data has been processed according to \cref{eq:aproc,eq:wproc}.
\end{itemize}

We provide two kinds of calibrated bags: one with full resolution and one with quarter resolution (half resolution for each dimension). The downsampled version facilitates usage for users with storage or bandwidth limitations.

\subsubsection{Calibration Files}
We also provide geometric calibration files which have been obtained from the processed calibration bags using the Kalibr toolbox\footnote{\url{https://github.com/ethz-asl/kalibr}} \cite{furgale2013unified}. They include intrinsic camera parameters for different models and the relative poses between cameras and IMU.
Additionally, the vignette calibration result is given for each camera in PNG format as described in \cref{sec:photocalib}.

\section{EVALUATION}

\begin{figure*}
\centering
\includegraphics[width=0.98\linewidth]{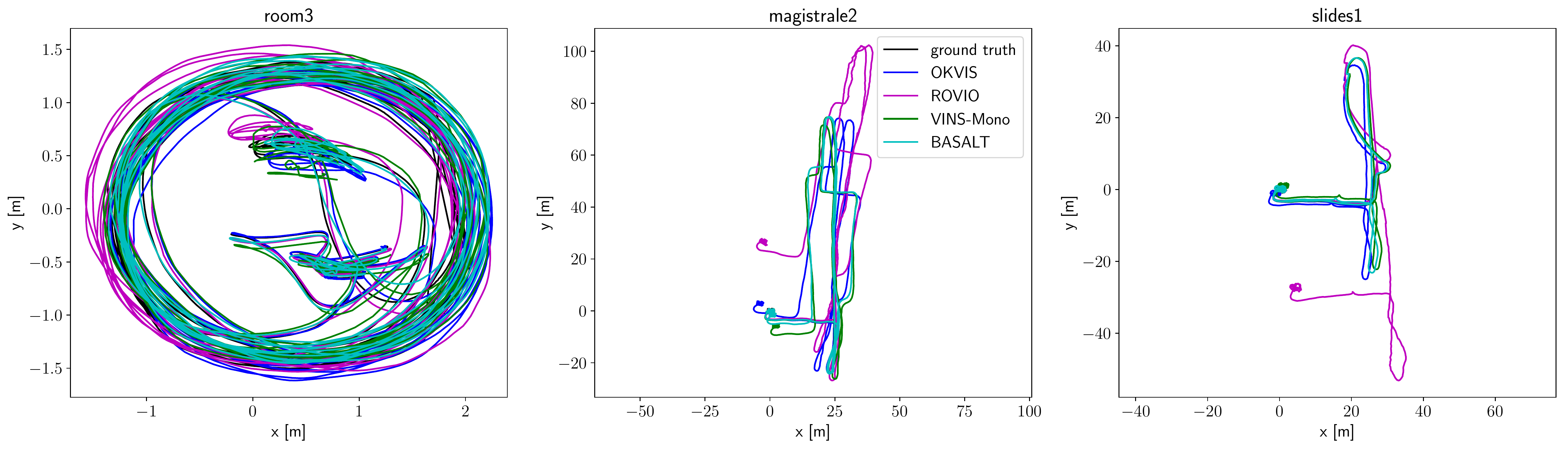}

\caption{Results of evaluated methods for room3, magistrale2 and slides1 sequences from our dataset. The ground truth is shown in black for the segments of the trajectory where it is available. The presented results are obtained with synchronous processing, without enforcing real-time and otherwise default parameters (except VINS-Mono for which non-realtime version is not implemented). Noise parameters are set to inflated values from the Allan plots in \cref{fig:allan} to account for unmodeled noise and vibrations.
}
\label{fig:eval_results}
\end{figure*}

\subsection{Evaluation Metric}

To evaluate the performance of tracking algorithms on the dataset, we use different evaluation metrics. The \emph{absolute trajectory error} is used, which is the root mean squared difference of ground-truth 3D positions $\hat{\mathbf{p}}_i$ and the corresponding tracked positions $\mathbf{p}_i$, aligned with an optimal $\mathrm{SE}(3)$ pose $\mathbf{T}$,

\begin{align}
    r_\text{ate} = \min_{\mathbf{T}\in \SE(3)} \sqrt{\frac{1}{|I_\text{gt}|}\sum_{i\in I_\text{gt}} \left\|\mathbf{T}\mathbf{p}_i-\hat{\mathbf{p}}_i\right\|^2}\,.
\end{align}
All tracked poses where ground truth is available are used, which corresponds to indices $I_\text{gt}$. For most sequences, this is the case at the start and at the end, but for some sequences, there is ground truth throughout.

For visual odometry without global optimization, another reasonable quantity is the \emph{relative pose error}. Following \cite{sturm2012benchmark}, it is defined as
\begin{align}
    r_\text{rpe} &= \sqrt{\frac{1}{|I_{\text{gt},\Delta}|}\sum_{i\in I_{\text{gt},\Delta}}\left\|\trans(\mathbf{E}_i)\right\|^2}\,,\\
    \mathbf{E}_i &= \left(\hat{\mathbf{T}}_i^{-1}\hat{\mathbf{T}}_{i+\Delta}\right)^{-1}\left(\mathbf{T}_i^{-1}\mathbf{T}_{i+\Delta}\right)\,,
\end{align}
where $\trans(\cdot)$ takes the 3D translational component of a pose. This error measures how accurate pose changes are in a small time interval $\Delta$. The set of frame indices $I_{\text{gt},\Delta}$ is the same as $I_{\text{gt}}$, but we have to take out $\Delta$ poses at the end of each tracked segment.

\subsection{Results}

To verify that the dataset is suitable for benchmarking visual-inertial odometry systems, we provide the results of several state-of-the-art methods that have open-source implementations. Unless specified otherwise, the methods are used with default parameters on quarter resolution images (512x512 pixels). We found that most of the algorithms have default parameters tuned to images with VGA resolution, which makes their performance better on sub-sampled datasets, while full resolution data might be useful for future research.

We provide evaluations for ROVIO \cite{bloesch17}, OKVIS \cite{leutenegger15}, VINS-Mono \cite{qin2017vins} and BASALT \cite{usenko_nfr}. The results are summarised in \cref{tab:rpe_results} and \cref{tab:ate_results} and a visualization for some sequences is presented in \cref{fig:eval_results}. All systems are able to track most of the sequences until the end, surprisingly, even the sequences with complete absence of visual features for some parts of the trajectory (slides). However, sometimes the estimators diverge at some point during the sequence, which results in erratic translation or rapid drift. We call a sequence diverged, if the ATE based on just the end-segment is larger\footnote{For all evaluated systems, median values over all sequences for ATE based on just the start-segment are less than \SI{0.1}{m} and less than \SI{0.5}{m} for just the end-segment.} than \SI{2}{m}, which is indicated by underlines in \cref{tab:ate_results}. The ATE values are still informative, as most often divergence happens towards the end (values larger than \SI{1000}{m} are shown as ``X'').

OKVIS, VINS-Mono and BASALT perform mostly well, but struggle for some of the longer outdoor sequences. ROVIO is more prone to drift and diverges on several sequences, which might be explained by it's use of a Kalman filter compared to computationally more demanding non-linear least squares optimization employed by OKVIS, VINS-Mono and BASALT. VINS-Mono diverges on most of the outdoor sequences, but typically only after the camera returns to the motion capture room and switches from mainly forward motion to fast rotations. This might indicate a drift in accelerometer bias estimates. 

The evaluation shows that even the best performing algorithms have significant drift in long (magistrale, outdoors) and visually challenging (slides) sequences. This means that the dataset is challenging enough to be used as a benchmark for further research in visual-inertial odometry algorithms.

\begin{table}
    \caption{RMSE ATE IN M OF THE EVALUATED METHODS}
    \label{tab:ate_results}
    \centering
    \begin{tabular}{lrrrrr}
        \toprule        
        Sequence & OKVIS & ROVIO & VINS & BASALT & length [m]\\
        \midrule
        
        corridor1 & \textbf{0.33} & 0.47 & 0.63 & 0.34 & 305 \\
        corridor2 & 0.47 & 0.75 & 0.95 & \textbf{0.42} & 322 \\
        corridor3 & 0.57 & 0.85 & 1.56 & \textbf{0.35} & 300 \\
        corridor4 & 0.26 & \textbf{0.13} & 0.25 & 0.21 & 114 \\
        corridor5 & 0.39 & 2.09 & 0.77 & \textbf{0.37} & 270 \\
        magistrale1 & 3.49 & 4.52 & 2.19 & \textbf{1.20} & 918 \\
        magistrale2 & 2.73 & 13.43 & 3.11 & \textbf{1.11} & 561 \\
        magistrale3 & 1.22 & 14.80 & \textbf{0.40} & 0.74 & 566 \\
        magistrale4 & \textbf{0.77} & 39.73 & 5.12 & 1.58 & 688 \\
        magistrale5 & 1.62 & 3.47 & 0.85 & \textbf{0.60} & 458 \\
        magistrale6 & 3.91 & X & \textbf{2.29} & 3.23 & 771 \\
        outdoors1 & X & 101.95 & \textbf{74.96} & 255.04 & 2656 \\
        outdoors2 & 73.86 & \textbf{21.67} & \underline{133.46} & 64.61 & 1601 \\
        outdoors3 & 32.38 & \textbf{26.10} & \underline{36.99} & 38.26 & 1531 \\
        outdoors4 & 19.51 & X & \underline{\textbf{16.46}} & 17.53 & 928 \\
        outdoors5 & 13.12 & 54.32 & \underline{130.63} & \textbf{7.89} & 1168 \\
        outdoors6 & 96.51 & 149.14 & \underline{133.60} & \textbf{65.50} & 2045 \\
        outdoors7 & 13.61 & 49.01 & 21.90 & \textbf{4.07} & 1748 \\
        outdoors8 & 16.31 & 36.03 & 83.36 & \textbf{13.53} & 986 \\
        room1 & \textbf{0.06} & 0.16 & 0.07 & 0.09 & 146 \\
        room2 & 0.11 & 0.33 & \textbf{0.07} & 0.07 & 142 \\
        room3 & \textbf{0.07} & 0.15 & 0.11 & 0.13 & 135 \\
        room4 & \textbf{0.03} & 0.09 & 0.04 & 0.05 & 68 \\
        room5 & \textbf{0.07} & 0.12 & 0.20 & 0.13 & 131 \\
        room6 & 0.04 & 0.05 & 0.08 & \textbf{0.02} & 67 \\
        slides1 & 0.86 & 13.73 & 0.68 & \textbf{0.32} & 289 \\
        slides2 & 2.15 & 0.81 & 0.84 & \textbf{0.32} & 299 \\
        slides3 & 2.58 & 4.68 & \textbf{0.69} & 0.89 & 383 \\


        \bottomrule
    \end{tabular}
\end{table}

\section{CONCLUSION}

In this paper, we proposed a novel dataset with a diverse set of sequences in different scenes for evaluating visual-inertial odometry. It contains high resolution images with high dynamic range and vignette calibration, hardware synchronized with 3-axis accelerometer and gyro measurements. For evaluation, the dataset contains accurate pose ground truth at high frequency at the start and end of the sequences. We perform hand-eye calibration on calibration sequences and time-offset estimation on all sequences to have ground truth data geometrically and temporally aligned with the IMU. In addition, we provide sequences to calibrate IMU white noise and random walk and vignetting of the camera. The dataset is publicly available with raw and calibrated data.

We also use our benchmark to evaluate the performance of state-of-the-art monocular and stereo visual-inertial methods. Our results demonstrate several open challenges for such approaches. Hence, our benchmark can be useful for the research community for evaluating visual-inertial odometry approaches in future research.

\section*{ACKNOWLEDGMENT}
This work was partially supported through the grant ``For3D" by the Bavarian Research Foundation and through the grant CR~250/9-2 ``Mapping on Demand" by the German Research Foundation.





\appendix

\subsection{Allan deviation for different noise types}\label{ap}

\subsubsection{Allan variance}

For a series of IMU measurements $g_i$ with time spacing $\tau_0$, we define an average over $n$ consecutive measurements starting at $g_i$ as
\begin{align}\label{eq:avgyr}
\bar{g}_i=\frac{1}{n}\sum_{j=i}^{i+n-1}g_j\,.
\end{align}
The Allan variance is defined as
\begin{align}\label{eq:allandef}
    \sigma_\text{A}^2(n\tau_0)=\frac{1}{2}\langle (\bar{g}_{i+n}-\bar{g}_i)^2\rangle\,,
\end{align}
which we estimate as
\begin{align}
    \hat\sigma_\text{A}^2(n\tau_0) = \frac{1}{2(M-2n+1)}\sum_{k=1}^{M-2n+1}(\bar{g}_{i+n}-\bar{g}_i)^2\,,
\end{align}
where $M$ is the total number of measurements.

\subsubsection{White noise on the measurements}
Consider the case where each measurement is perturbed by independent Gaussian noise $\epsilon_i\sim\mathcal{N}(0,\sigma^2_{\text{w},\tau_0})$. For a resting IMU, the measurements are not necessarily centered around zero, due to a possible bias plus gravity in the case of the accelerometer, so the measurements are modeled as
\begin{align}
    g_i=\mu+\epsilon_i\,.
\end{align}
Plugging this into Eqs. \ref{eq:avgyr} and \ref{eq:allandef}, one obtains
\begin{align}
   \sigma_\text{A}^2(n\tau_0) &= \frac{1}{2}\langle (\bar{\epsilon}_{i+n}-\bar{\epsilon}_i)^2\rangle\\
                            &= \frac{1}{2}\left(\langle\bar{\epsilon}_i^2\rangle+\langle\bar{\epsilon}_{i+n}^2\rangle-2\langle\bar{\epsilon}_i\bar{\epsilon}_{i+n}\rangle\right)\label{eq:allanexpand}\\
                            &= \langle\bar{\epsilon}_i^2\rangle\\
                            &= \frac{\sigma_{\text{w},\tau_0}^2}{n} \,.\label{eq:allanvar}
\end{align}
The average of the error $\bar{\epsilon}_i$ is defined analogously to $\bar{g}_i$. In Eq.~\ref{eq:allanexpand}, the term $\langle\bar{\epsilon}_i\bar{\epsilon}_{i+n}\rangle$ vanishes, as both $\bar{\epsilon}_i$ and $\bar{\epsilon}_{i+n}$ have zero mean and are uncorrelated. Also, $\langle\bar{\epsilon}_i\rangle$ and $\langle\bar{\epsilon}_{i+n}\rangle$ are equal, as they are calculated from identically distributed random variables. Taking the square root of Eq.~\ref{eq:allanvar} and substituting $n$ by $\tau/\tau_0$ results in the Allan deviation
\begin{align}
\sigma_\text{A}(\tau) &= \frac{\sigma_{\text{w},\tau_0}\sqrt{\tau_0}}{\sqrt{\tau}}\\
                    &=\frac{\sigma_\text{w}}{\sqrt{\tau}}\,,\label{eq:allandev}
\end{align}
where $\sigma_\text{w}=\sigma_{\text{w},\tau_0}\sqrt{\tau_0}$ is the continuous-time parameter. From Eq.~\ref{eq:allandev} one can deduct that
\begin{align}
    \sigma_\text{w}= \sigma_\text{A}(\SI{1}{\second})\sqrt{\si{\second}}\,.
\end{align}
Note that $\si{\second}$ is a unit here, not a variable. In order to obtain an estimate for $\sigma_\text{A}(\SI{1}{\second})$ and thus for $\sigma_\text{w}$, a function can be fitted to $\sigma_\text{A}(\tau)$ and evaluated at $\tau=\SI{1}{\second}$. As $\sigma_\text{A}(\tau)$ in Eq.~\ref{eq:allandev} is a power function with exponent $-1/2$, it will appear as a straight line with slope $-1/2$ in a log-log plot, which means the fit only includes determining the offset of a straight line with given slope.

\subsubsection{Bias random walk}

Now the errors $\epsilon_i = \sum_{j=1}^i X_j$ are a summation of Gaussian variables $X_j\sim\mathcal{N}(0,\sigma^2_{\text{b},\tau_0})$.
This means that
\begin{align}
    \langle\epsilon_i\epsilon_j\rangle &= \sum_{k=1}^i\sum_{l=1}^j \langle X_k X_l\rangle\\
                                        &= \sum_{k=1}^{\min(i,j)} \langle X_k^2\rangle\\
                                        &= \min(i,j)\sigma^2_{\text{b},\tau_0}\,.
\end{align}

Continuing from Eq.~\ref{eq:allanexpand}, one obtains
\begin{align}
    \sigma_\text{A}^2(n\tau_0) &= \frac{1}{2 n^2} \left(\sum_{j=i}^{i+n-1}\sum_{k=i}^{i+n-1}\langle\epsilon_j\epsilon_k\rangle\right.\\
                                                        &\qquad+\sum_{j=i+n}^{i+2n-1}\sum_{k=i+n}^{i+2n-1}\langle\epsilon_j\epsilon_k\rangle\\
                                                    &\qquad\left.-2\sum_{j=i}^{i+n-1}\sum_{k=i+n}^{i+2n-1}\langle\epsilon_j\epsilon_k\rangle\right)\\
                                &= \frac{\sigma^2_{\text{b},\tau_0}}{2 n^2} \left( \sum_{j=i}^{i+n-1}\sum_{k=i}^{i+n-1}\min(j,k)\right.\\
                                                    & \qquad   + \sum_{j=i+n}^{i+2n-1}\sum_{k=i+n}^{i+2n-1}\min(j,k)\\
                                                    &  \qquad  \left.- 2\sum_{j=i}^{i+n-1}\sum_{k=i+n}^{i+2n-1}\min(j,k)\right)\\
                                &= \sigma^2_{\text{b},\tau_0}\left( \frac{1}{6n} + \frac{n}{3}\right)\,.
\end{align}
The last step was done with the help of a computer algebra software. Taking only the leading power of $n$, the Allan deviation for large $n$ becomes
\begin{align}
    \sigma_\text{A}(\tau) &\approx \frac{ \sigma_{\text{b},\tau_0}}{\sqrt{\tau_0}}\sqrt{\frac{\tau}{3}}\\
                    &=\sigma_\text{b}\sqrt{\frac{\tau}{3}}\,,\label{eq:allandevbias}
\end{align}
with $\sigma_\text{b} = \sigma_{\text{b},\tau_0}/\sqrt{\tau_0}$. This means
\begin{align}
\sigma_\text{b} = \sigma_\text{A}(\SI{3}{\second})/\sqrt{\si{\second}}\,,
\end{align}
and this time the slope in the log-log plot is $+1/2$.

\balance

\bibliographystyle{IEEEtran}
\bibliography{IEEEabrv,IEEEexample}

\end{document}